%% file: iclr2023_conference.tex
\documentclass{article} 
\usepackage[final]{neurips}

\input{math_commands.tex}

\usepackage{hyperref}
\usepackage{url}
\usepackage{graphicx}
\usepackage{wrapfig}
\usepackage{booktabs}

\usepackage{algorithm}
\usepackage{algpseudocode}

\hypersetup{
  colorlinks,
  citecolor=blue,
  linkcolor=black,
  urlcolor=blue}

\title{\textit{Capsa}: A Unified Framework for Quantifying Risk in Deep Neural Networks}

\author{Sadhana Lolla\thanks{Denotes equal contribution and co-first authorship}~, Iaroslav Elistratov$^*$, Alejandro Perez, Elaheh Ahmadi, \\\textbf{Daniela Rus, Alexander Amini\thanks{Corresponding author: \texttt{alexander@themisai.io}}}
\\
Themis AI Inc\\
\url{themisai.io}\\
}

\def\capsa{\texttt{{capsa}}}
\def\Capsa{\texttt{{Capsa}}}

\begin{document}

\maketitle

\begin{abstract}
The modern pervasiveness of large-scale deep neural networks (NNs) is driven by their extraordinary performance on complex problems but is also plagued by their sudden, unexpected, and often catastrophic failures, particularly on challenging scenarios. 
%
Existing algorithms that provide risk-awareness to NNs are complex and ad-hoc. Specifically, these methods require significant engineering changes, are often developed only for particular settings, and are not easily composable. 
%
Here we present \capsa, a framework for extending models with risk-awareness. \Capsa{} provides a methodology for quantifying multiple forms of risk and composing different algorithms together to quantify different risk metrics in parallel. 
%
We validate \capsa{} by implementing state-of-the-art uncertainty estimation algorithms within the \capsa{} framework and benchmarking them on complex perception datasets. We demonstrate \capsa{}'s ability to easily compose aleatoric uncertainty, epistemic uncertainty, and bias estimation together in a single procedure, and show how this approach provides a comprehensive awareness of NN risk. 
\end{abstract}

\section{Introduction}



Neural networks (NNs) continue to push the boundaries of modern artificial intelligence (AI) systems across a wide range of complex real-world domains, from robotics and autonomy~\citep{bojarski2016end, hawke2020urban, codevilla2018end}, to healthcare and medical decision making~\citep{ching2018opportunities, topol2019high}. While their performance in these domains remains unmatched, modern NNs still encounter sudden, unexpected, and inexplicable failures that are often catastrophic -- especially in safety-critical environments. These failures are largely due to systemic issues that propagate throughout the entire modern AI lifecycle, from imbalances \citep{he2009learning, buda2018systematic} and noise \citep{beigman2009learning} in data that lead to algorithmic bias \citep{bolukbasi2016man,caliskan2017semantics, buolamwini2018gender,chen2018my, obermeyer2019dissecting, seyyed2021underdiagnosis} to predictive uncertainty \citep{kendall2017uncertainties, kompa2021second, nado2021uncertainty, amini2020deep} that plagues model performance on unseen or out-of-distribution data. In order to realize the widespread adoption of AI in society, NN models must not only identify these potential failure modes, but also effectively use this awareness to obtain unified and calibrated measures of risk and uncertainty. There is thus a critical need for unified systems that can estimate quantitative risk metrics for any NN model, and in turn integrate this awareness back into the learning lifecycle to improve robustness, generalization, and safety.

\begin{figure*}[t!]
\vspace{-10pt}
\centering
\includegraphics[width=1\linewidth]{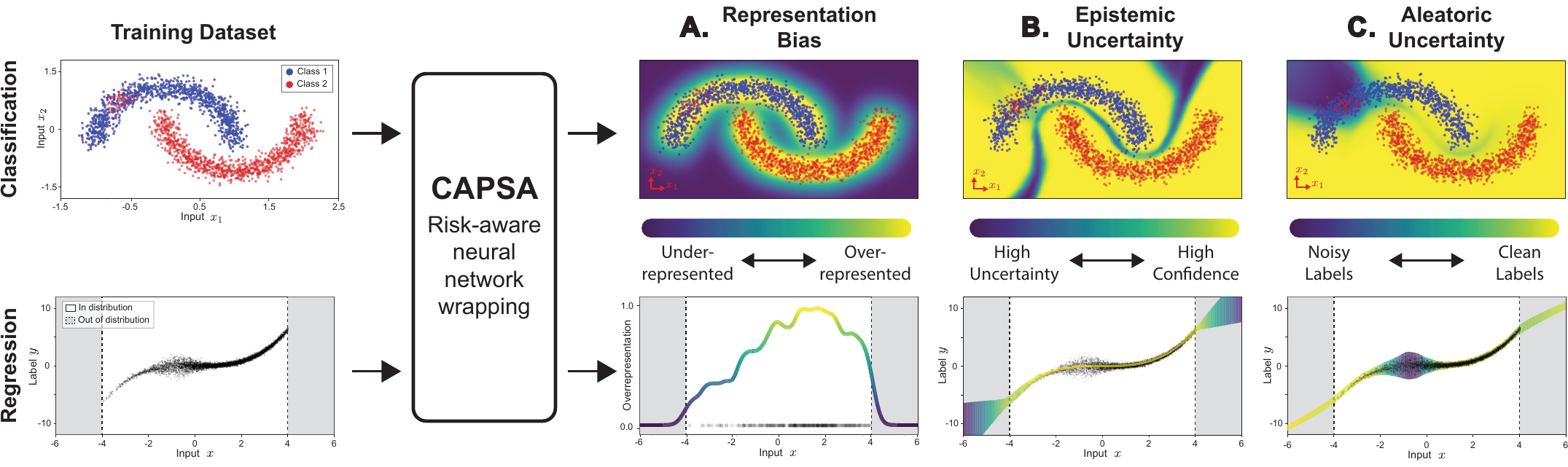}
\caption{\textbf{\Capsa}~ unifies state-of-the-art algorithms for quantifying neural network risks ranging from (A) under-representation bias, (B) epistemic (model) uncertainty, and (C) aleatoric uncertainty (label noise). \Capsa~ converts existing models into risk-aware variants, capable of identifying risks efficiently during training and deployment. }
\vspace{-10pt}
\end{figure*}

Existing algorithmic approaches to risk quantification narrowly estimate a singular form of risk in AI models, often in the context of a limited number of data modalities~\citep{nix1994estimating, kendall2017uncertainties, lakshminarayanan2017simple,buolamwini2018gender, zhang2018mitigating, gilitschenski2019deep}. These methods present critical limitations as a result of their reductionist, ad hoc, and narrow focus on single metrics of risk or uncertainty. However, generalizable methods that provide a larger holistic awareness of risk have yet to be realized and deployed~\citep{nado2021uncertainty, tran2022plex}. This is in part due to the significant engineering changes required to integrate an individual risk algorithm into a larger machine learning system~\citep{tran2016edward, dillon2017tensorflow, bingham2019pyro, shi2017zhusuan}, which in turn can impact the quality and reproducibility of results. The lack of a unified approach for composing different risk estimation algorithms or risk-aware models limits the scope and capability of each algorithm independently, and further limits the robustness of the system as a whole. A general, model-agnostic framework for extending NN systems with holistic risk-awareness, covering both uncertainty and bias, would advance the ability and robustness of end-to-end systems.

To address these fundamental challenges, we present \capsa{} -- an algorithmic framework for wrapping any arbitrary NN model with state-of-the-art risk-awareness capabilities. By decomposing the algorithmic stages of risk estimation into their core building blocks, we unify different algorithms and estimation metrics under a common data-centric paradigm. Additionally, because \capsa{} allows the underlying NN to be aware of a variety of risk metrics in parallel, we achieve improved performance and quality in risk estimation through principled redundancy, and open the door to achieving a unified composition and hierarchical understanding of NN risk.

In summary, the key contributions of this paper are:
\begin{enumerate}
    \item \Capsa{}, a flexible, and easy-to-use framework for equipping any given neural network with calibrated awareness of different forms of risk -- including bias, label noise, and predictive uncertainty;  
    \item An algorithm for decomposing different types of risk and their estimation methods into modular components that can in turn be integrated and composed together to achieve greater accuracy, robustness, and efficiency; and
    \item Empirical validation of \capsa{} on a range of dataset complexities and modalities, along with the application of \capsa{} for mitigation of algorithmic bias, identification of label noise, and detection of anomalies and out-of-distribution data.
\end{enumerate}

We refer readers to \texttt{{Capsa Pro}} \citep{capsa-pro} for information on the software library with the full functionality described in this publication.





\section{Background and Methodology}
\begin{figure*}[b!]
\vspace{-10pt}
\centering
\includegraphics[width=\linewidth]{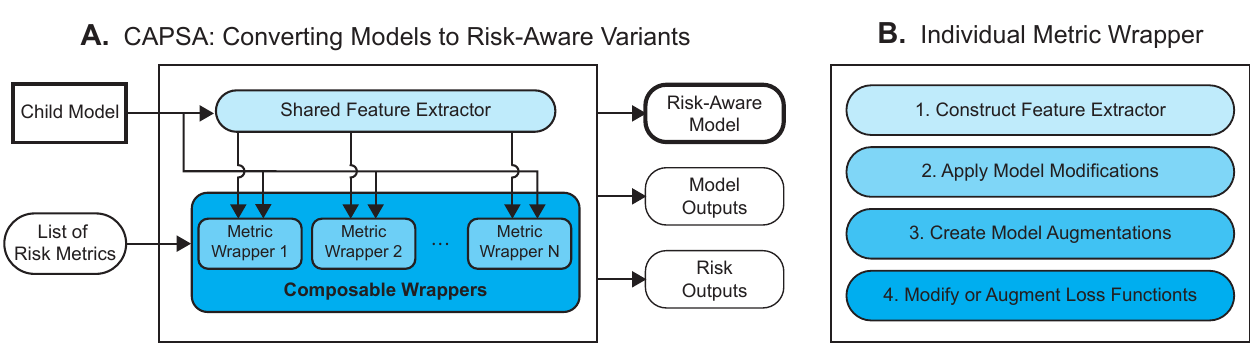}
\caption{\textbf{Overview of \Capsa~ architecture}. (A) \Capsa~ converts arbitrary NN models into risk-aware variants, that can simultaneously predict both their output along with a list of user-specified risk metrics. (B) Each risk metric forms the basis of a singular model wrapper which is constructed through metric-specific modifications to the model architecture and loss function.}
\label{fig:teaser}
\end{figure*}

\subsection{Preliminaries}

We consider the problem of supervised learning, where we are given a labeled dataset of $n$ input, output pairs, $\{x, y\}_{i=1}^n$. Our goal is to learn a model, $f$, parameterized by weights, $\bm{W}$, that minimizes the average loss over the entire dataset: $\sum_i \mathcal{L}(f_{\bm{W}}(x), y)$. While traditionally, the model, a neural network, outputs predictions in the form of $\hat y = f_{\bm{W}}(x)$, we now introduce a risk-aware transformation operation, $\Phi$, which transforms a model, $f$, into a risk-aware variant, such that
\begin{align*}
g = \Phi_{\bm \theta}(f_{\bm W}),\\
\hat y, R = g (x),
\end{align*}
%
%
where $R$ are the estimated ``risk'' measures from a set of metrics, $\bm \theta$. The goal of this paper is to propose a common transformation backbone for $\Phi_{\bm \theta}(\cdot)$, which automatically transforms an arbitrary model, $f$, to be aware of risks, $\bm \theta$.

All measures of risk aim to capture, on some level, the reliability of a given prediction. This can stem from the data source (aleatoric uncertainty, or representation bias) or the predictive capacity of the model itself (epistemic uncertainty). Within \capsa, we define various risk metrics to identify and measure these sources of risk. We propose the idea of \textit{wrappers}, which are instantiations of $\Phi_\theta$, for a singular risk metric, $\theta$. Wrappers are given an arbitrary neural network and, while preserving the structure and function of the network, add and modify the relevant components in the model. This allows them to serve as a drop-in replacement that is able to estimate the risk metric, $\theta$. Wrappers can be further composed using a set of metrics, $\bm \theta$, that are faster and more accurate than individual metrics. 

\subsection{Capsa: The Wrapping Algorithm}

While risk estimation algorithms can take a variety of forms and are often developed in ad hoc settings, we present a unified algorithm for building $\Phi_{\bm \theta}$ in order to wrap an arbitrary neural network model. There are four main components: (1) constructing the shared feature extractor, (2) applying modifications to the existing model needed to capture the uncertainty, (3) creating additional models and augmentations if necessary, and (4) modifying the loss functions. 

The feature extractor, which defaults to the model until its last layer, can be leveraged as a shared backbone by multiple wrappers at once to predict multiple compositions of risk. This results in a fast, efficient method of reusing the main body of the model, which does not require training multiple models and risk estimation methods from scratch. Next, \capsa~ modifies the existing network according to metric-specific modifications; for example, this could entail modifying every weight in the model to be drawn from a distribution (to convert to a Bayesian neural network~\citep{blundell2015weight}) or adding stochastic dropout layers~\citep{gal2016dropout}. Depending on the metric, \capsa~ also adds new layers or augmentations to the model that predict new outputs. Note that these are not modifications to the model, but rather augmentations for the given metric; for example, new layers to output $\sigma$~\citep{nix1994estimating}, or extra model copies when ensembling~\citep{lakshminarayanan2017simple}. Lastly, we modify the loss function to capture any remaining metric-specific changes that need to be made. This entails combining the user-specified and metric-specific loss functions (e.g., KL-divergence~\citep{kingma2013auto}, negative log-likelihood~\citep{nix1994estimating}, etc). All of the following modifications are integrated together into a custom metric-specific forward pass, and train step. These are used to capture variations in the forward and backward passes during training and inference. 

\subsection{Risk Metrics and Background}
In this section, we outline three high-level categories of risk which we quantitatively define and estimate.

\textbf{Representation Bias - } 
The representation bias of a dataset uncovers imbalance in the space of features and captures whether certain combinations are more prevalent than others. Note that this is fundamentally different from label imbalance, which only captures distributional imbalance in the labels. For example, in driving datasets, it has been demonstrated that the combination of straight roads, sunlight, and absence of traffic is higher than any other feature combinations. This  indicates that these samples are overrepresented~\citep{amini2018variational}. Similar combinations have  been identified for facial detection~\citep{buolamwini2018gender, amini2019uncovering},  medicine~\citep{puyol2021fairness, soleimany2021evidential}, and clinical trials~\citep{xu2022identifying}. Uncovering feature representation bias is a computationally expensive process as these features are (1) often unlabeled, and (2) extremely high-dimensional (e.g., images, videos, language, etc). However, they can be estimated by learning the density distribution of the data. We accomplish this by estimating densities in feature space. For high-dimensional feature spaces we estimate a low-dimensional embedding using a variational autoencoder~\citep{kingma2013auto} or by using the features from the penultimate layer of the model. Bias is then the imbalance between parts of the density space estimated either discretely (using a discretely-binned histogram) or continuously (using a kernel distribution~\citep{rosenblatt1956remarks}). 

\begin{wrapfigure}[10]{R}{0.5\textwidth}
\begin{minipage}{0.48\textwidth}
\vspace{-20pt}
\begin{algorithm}[H]
    \centering
    \small
    \caption{\small{Aleatoric Uncertainty in Classification}}
    \label{alg:class-mve}
    \begin{algorithmic}[1]
        \State $\mu, \sigma \leftarrow f_{\bm{W}}(x)$ \Comment{Inference}
        \For{$i \in 1..T$} \Comment{Stochastic logits}
        \State $\tilde{z} \leftarrow \mu + \sigma \times \epsilon \sim \mathcal{N}(0, 1)$
        \EndFor
        \State $\tilde{z} \leftarrow \frac{1}{N} \times \sum_{i = 1}^{T} \tilde{z}$ \Comment{Average logit}
        \State $\hat y \leftarrow \frac{\exp(\tilde{z})}{\sum_j \exp(\tilde{z}_j)}$ \Comment{Softmax probability}
        \State $\mathcal{L}(x, y) \leftarrow -\sum_j y_j \log p_j$ \Comment{Cross entropy loss}
    \end{algorithmic}
\end{algorithm}
\end{minipage}
\end{wrapfigure}

\textbf{Aleatoric Uncertainty- }  
Aleatoric uncertainty captures noise in the data, e.g., mislabeled datapoints, ambiguous labels, classes with low separation, etc. We model aleatoric uncertainty using Mean and Variance Estimation (MVE)~\citep{nix1994estimating}. In the regression case, we pass the outputs of the model's feature extractor to another layer that predicts the standard deviation of the output. We train using NLL, and use the predicted variance as an estimate of the aleatoric uncertainty. We apply a modification to the algorithm to generalize to the classification case in ~\alg{alg:class-mve}. We assume the classification logits are drawn from a normal distribution and stochastically sample from them using the reparametrization strategy. We average stochastic samples and backpropogate using cross entropy loss through logits and their inferred uncertainties.

\textbf{Epistemic Uncertainty- }
Epistemic uncertainty measures uncertainty in the model's predictive process -- this captures scenarios such as examples that are "hard" to learn, examples whose features are underrepresented, and out-of-distribution data. We provide a unified approach for a variety of epistemic uncertainty methods ranging from Bayesian neural networks~\citep{blundell2015weight}, ensembling~\citep{lakshminarayanan2017simple}, and reconstruction-based~\citep{kingma2013auto} approaches. Below, we outline three metrics and how they fit into \capsa's unified risk estimation framework.  

A \textit{Bayesian neural network} can be approximated by stochastically sampling, during inference, from a neural network with probabilistic layers~\citep{blundell2015weight, gal2016dropout}. Adding dropout layers~\citep{srivastava2014dropout} to a model is one of the simplest ways to capture epistemic uncertainty~\citep{gal2016dropout}. To calculate the uncertainty, we run $T$ forward passes, which is equivalent to Monte Carlo sampling. Computing the first and second moments from the $T$ stochastic samples yields a prediction and uncertainty estimate, respectively. 

An \textit{ensemble} of $N$ models, each a randomly initialized stochastic sample, is a common approach used to accurately estimate epistemic uncertainty~\citep{lakshminarayanan2017simple}. However, this comes with significant computational costs. To reduce the cost of training ensembles, \capsa~ automates the construction and management of the training procedure for all members and parallelizes their computation. 

\textit{Variational autoencoders (VAEs)} are typically used to learn a robust, low-dimensional representation of the latent space. They can be used as to estimate epistemic uncertainty by using the reconstruction loss $MSE(\hat{x}, x)$. In cases of out-of-distribution data, samples that are hard to learn, or underrepresented samples, we expect that the VAE will have high reconstruction loss, since the mapping to the latent space will be less accurate. Conversely, when the model is very familiar with the features, or the data is in distribution, we expect the latent space mapping to be robust and the reconstruction loss to be low. To construct the VAE for any given model in \capsa, we use the feature extractor as the encoder, and reverse the feature extractor automatically when possible to create a decoder. 

\subsection{Metric Composability}

Using \capsa, we compose multiple risk metrics to create more robust estimates (e.g., by combining multiple metrics, or alternatively by capturing different measures of risk independently). By using the feature extractor as a shared common backbone, we can optimize for multiple objectives, ensemble multiple metrics, and obtain different types of uncertainty estimates simultaneously.

We propose a novel composability algorithm within \capsa~ to automate this process. Again, we leverage our shared feature extractor as the common backbone for all metrics and incorporate all model modifications. Then, we apply the new model augmentations either in series or in parallel, depending on the use case (i.e., we can ensemble a metric in series to average the metric over multiple joint trials, or we can apply ensembling in parallel to estimate a independent measure of risk). Lastly, the model is jointly optimized using all of the relevant loss functions by computing the gradient of each one with regard to the shared backbone’s weights and stepping into the direction of the accumulated gradient.


\section{Experimental Results}
In the following section, we analyze the risk metrics obtained by wrapping various models with \capsa~ on several datasets. We show that \capsa~ provides accurate, scalable, composable risk metrics that are efficient and can be used to quantify bias, aleatoric, and epistemic uncertainty using multiple methods. 
\begin{figure}[t!]
\centering
\includegraphics[width=\linewidth]{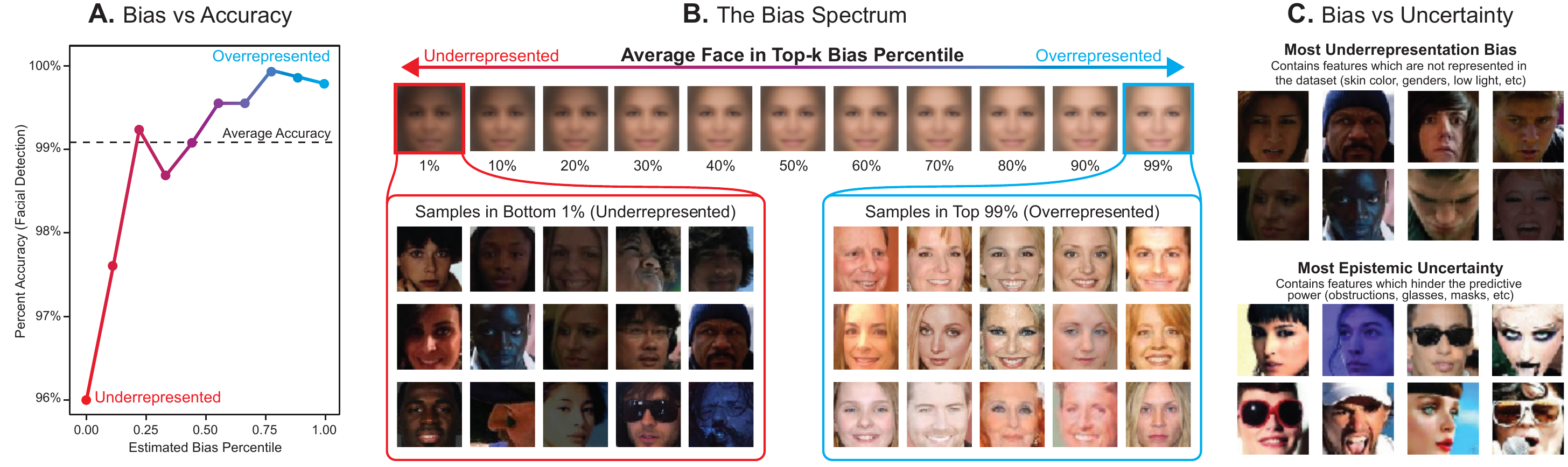}
\caption{\textbf{Bias and Epistemic Uncertainty on Faces} (A) Under-represented and over-represented faces in the Celeb-A dataset found by \capsa~ using the VAE and HistogramBias wrappers. As the percentile bias of the data increases, the skin tone gets lighter, lighting gets brighter, and hair color gets lighter, and (B) accuracy on these datapoints increases. We also determine the points with the highest epistemic uncertainty, which have artifacts such as sunglasses, hats, colored lighting, etc.}
\label{fig:top-bottom-k-bias}
\end{figure}

\subsection{Representation Bias}

Using \capsa's bias and epistemic wrapper capabilities, we analyzed the Celeb-A \citep{liu2015faceattributes} dataset. The task for the neural network was to detect faces from this dataset against non-face images (collated from various negatives in the ImageNet dataset).  \fig{fig:top-bottom-k-bias}A quantifies an accuracy vs bias tradeoff that neural networks exhibit, where they tend to perform better on overrepresented training features. We used a VAE as a feature extractor to demonstrate that the it can be used for single-shot bias and epistemic uncertainty estimation without any added computation cost.
    
\fig{fig:top-bottom-k-bias}B qualitatively inspects the different percentiles of bias ranging from underrepresentation (left) to overrepresentation (right). We found that the underrepresented samples in the dataset commonly contained darker skin tones, darker lighting, and faces not looking at the camera. As the percentile of the bias gets higher, we see that the dataset is biased towards lighter skin tones, hair colors, and a more uniform facial direction. With our approach, we highlight a critical difference between bias and epistemic estimation methods in \fig{fig:top-bottom-k-bias}C. The samples estimated to have the highest epistemic uncertainty were not necessarily only underrepresented, but also contain features that obscure the predictive power of the model (e.g., faces with colored lighting, covering masks, and artifacts such as sunglasses and hats).

\subsection{Aleatoric Uncertainty}
\label{res:aleatoric}

\begin{wrapfigure}[22]{R}{0.4\textwidth}
\centering
\vspace{-15pt}
\includegraphics[width=1\linewidth]{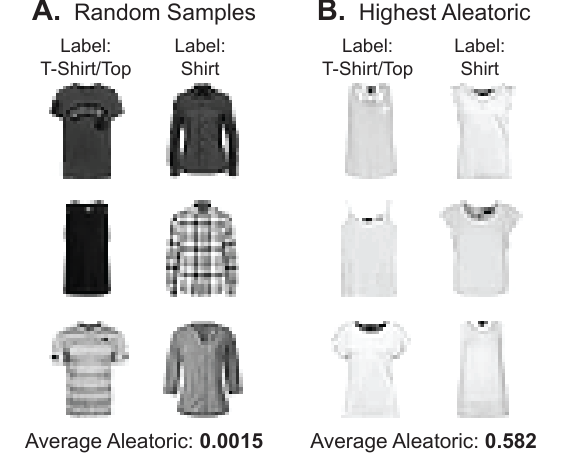}
\vspace{-10pt}
\caption{\textbf{Fashion MNIST Aleatoric Uncertainty} (A) Randomly selected samples from two classes of fashion-mnist. These samples are visually distinguishable, and have a low aleatoric uncertainty, as opposed to (B), which shows samples with highest estimated aleatoric noise. It is not clear what features distinguish these shirts from tshirts/tops, as they have similar necklines, sleeve lengths, and cuts.}
\label{fig:fashion-mnist-aleatoric}
\vspace{-40pt}
\end{wrapfigure}


Next, we experiment on \capsa's ability to successfully detects label noise in datasets using aleatoric uncertainty estimation. An example of this can be shown in Fashion-MNIST, which contains two very similar classes: ``tshirt/top'' and ``shirt''. The methods presented in \capsa~ identify samples in Fashion-MNIST with high aleatoric uncertainty, which are light sleeveless tops with similar necklines with minimal visual differences. Short-sleeved shirts with round necklines are also classified as either category. Compared to randomly selected samples from these two classes, the samples considered noisy by \capsa~ are visually indistinguishable, and difficult for humans (and models) to categorize. 

\subsection{Epistemic Uncertainty}

In this section, we benchmark \capsa's epistemic methods on toy datasets. We demonstrate how \capsa's ability to compose multiple methods (e.g.,  dropout and VAEs) can achieve more robust, efficient performance. We combine aleatoric methods with epistemic methods (i.e., ensembling the MVE metric) to strengthen aleatoric methods, since they are averaged across multiple runs. We can also treat the ensemble of MVEs as a mixture of normals. Similarly, to combine VAE and dropout, we use a weighted sum of their variances or we run the VAE $N$ times with dropout layers and treat multiple runs as $N$ normals.

\subsubsection{Cubic Dataset and UCI Benchmarking}

\begin{figure}[t!]
\centering
\includegraphics[width=1\linewidth]{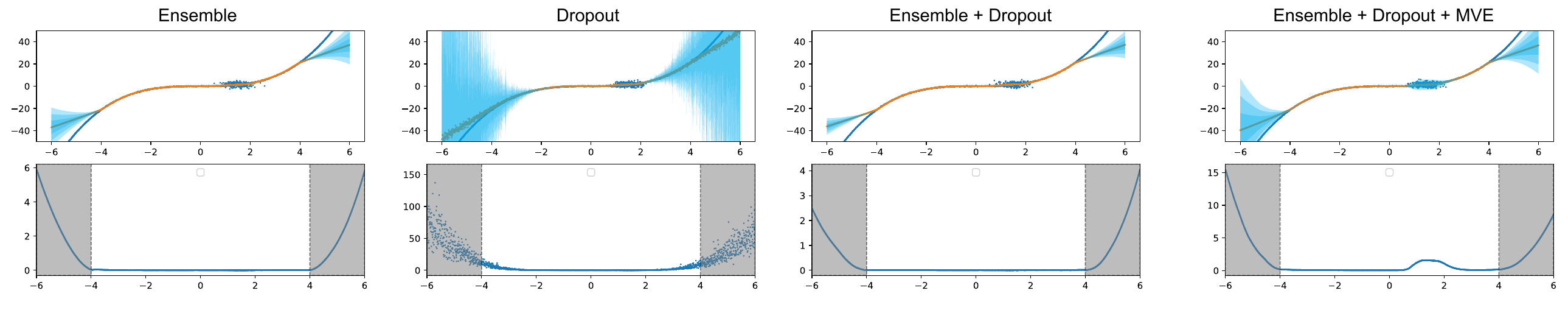}
\vspace{-15pt}
\caption{\textbf{Risk metrics on cubic regression.} A regression dataset $y = x + \epsilon$, where $\epsilon$ is drawn from a Normal centered at $x=1.5$. Models are trained on $x \in [-4,4]$ and tested on $x \in [-6,6]$. Composing using MVE results in a single metric that can seamlessly detect epistemic and aleatoric uncertainty without any modifications to the model construction or training procedure.}
\vspace{-10pt}
\label{fig:cubics}
\end{figure}

We compose various epistemic and aleatoric methods on a cubic dataset with injected aleatoric noise and a lack of data in some parts of the test set. We train models on $y = x + \epsilon$, where $\epsilon \sim \mathcal{N}(1.5, 0.9)$. Training data is within $[-4,4]$ and test within $[-6,6]$. \fig{fig:cubics} demonstrates that composed metrics can successfully detect regions with no data, as well as the aleatoric uncertainty in the center.

Additionally, we benchmark raw epistemic uncertainty methods on real-world regression datasets, and evaluate VAEs, ensembles, and dropout uncertainty on these datasets based on Root Mean Squared Error (RMSE) and negative log-likelihood (NLL) in ~\tab{tab:uci-benchmarks}. More composability results, as well as training times for all methods, are available in the appendix in ~\tab{tab:training-times} and ~\tab{tab:vae-dropout}. 
\input{uci_benchmark_table}

\subsubsection{Depth Estimation}

\begin{figure}[t!]
\centering
\includegraphics[width=\linewidth]{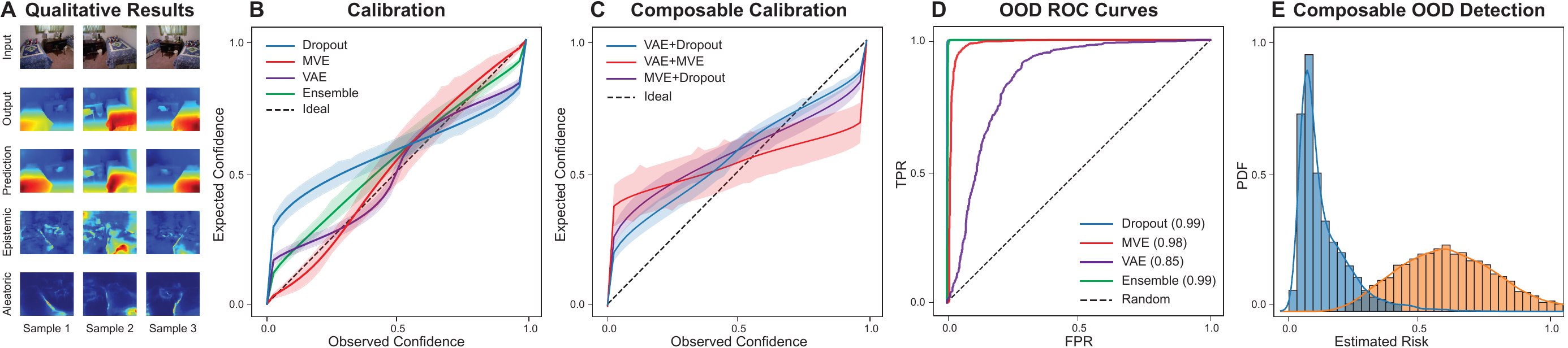}
\caption{\textbf{Risk estimation on monocular depth prediction}. (A) Example pixel-wise depth predictions and uncertainty. Model uncertainty calibration for individual metrics (B) and composed metrics (C). OOD detection assessed via AUC-ROC (D) and a full p.d.f. histogram (E).}
\label{fig:depth}
\end{figure}

In this section, we transition to more complex models and datasets and demonstrate how \capsa~ can be used as a large-scale risk and uncertainty benchmarking framework for existing methods. To that end, we train a U-Net style model on the task of monocular end-to-end depth estimation (see ~\tab{tab:depth-benchmark}). Importantly, \capsa~ works ``out of the box'' without requiring any modifications since it is a highly configurable, model-agnostic framework with modularity as one of the core of its design principles.

Specifically, we take a U-Net style model whose final layer outputs a single $H \times W$ activation map and wrap it with \capsa. We then train the wrapped model on NYU Depth V2 dataset~\citep{Silberman:ECCV12} (27k RGB-to-depth image pairs of indoor scenes) and evaluate on a disjoint test-set of scenes. Additionally, we use outdoor driving images from ApolloScapes~\citep{liao2020dvi} as OOD data points.

\input{depth_benchmark_table}

We see that when we wrap the model with an aleatoric method in \fig{fig:depth_mve},  we can successfully detect label noise or mislabeled data. The model exhibits increased aleatoric uncertainty on object boundaries. Indeed, we see that the ground truth has noisy labels particularly along the edges of objects which could be due to sensor noise or motion noise.

With dropout (\fig{fig:depth_dropout}) or ensemble (\fig{fig:depth_ensemble}) wrappers, we capture uncertainty in the model's prediction. We see that increased epistemic uncertainty roughly corresponds to the semantically and visually challenging pixels where the model returns erroneous output.

\section{Applications}
The benefits of seamlessly and efficiently integrating a variety of risk estimation methods into arbitrary neural models extends far beyond benchmarking and unifying these algorithms. In this section, we outline critically important applications that are possible with the estimation abilities in \capsa.

\subsection{Debiasing Facial Recognition Systems}

\begin{figure}[t!]
\centering
\includegraphics[width=1\linewidth]{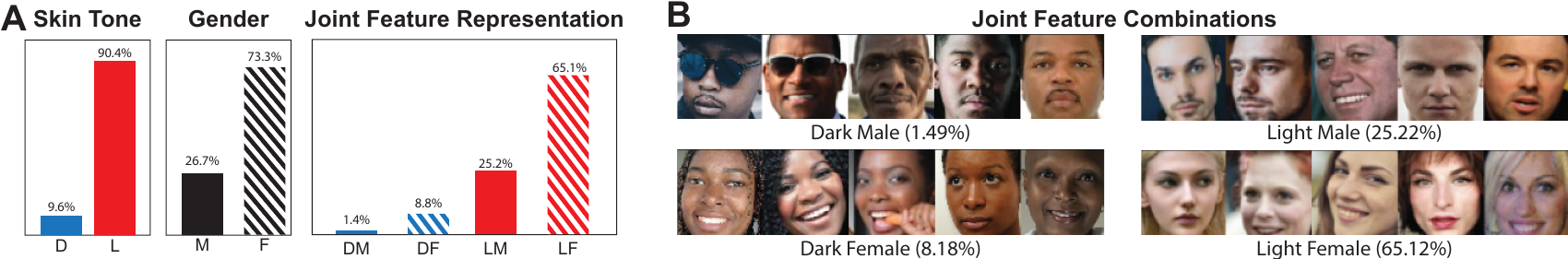}
\caption{\textbf{Debiasing Facial Recognition Systems} (A) Facial datasets are overwhelmingly biased towards light-skinned females. (B) The feature combinations present in dark-skinned males make up only 1.49\% of the dataset, and those present in dark-skinned female faces only take up 8.18\% of the dataset. Since \capsa{} identifies underrepresented datapoints, we can implement smart sampling schemes that increase the representation of these feature combinations.}
\label{fig:face-probabilities}
\end{figure}

Using the bias tools provided by \capsa, one application is to not only estimate and identify imbalance in the dataset (which we show also leads to performance bias) but to actively reduce the performance bias by adaptively re-sampling datapoints depending on their estimated representation bias during the course of training. As shown in \fig{fig:face-probabilities}, using \capsa~, we pinpoint exactly which samples need under/oversampling, and therefore can intelligently resample from the dataset during training. The benefits of this are twofold -- we can improve sample efficiency by training on less data if some data is redundant, and we can also oversample from areas of the dataset where our latent representation is more sparse. 

By composing multiple risk metrics together (in this case, VAEs and histogram bias) we can achieve even greater robustness during training, more sample efficiency, and combine epistemic uncertainty and bias to reduce risk.

\subsection{Detecting mislabeled examples}
\begin{figure*}[t!]
\centering
\includegraphics[width=\linewidth]{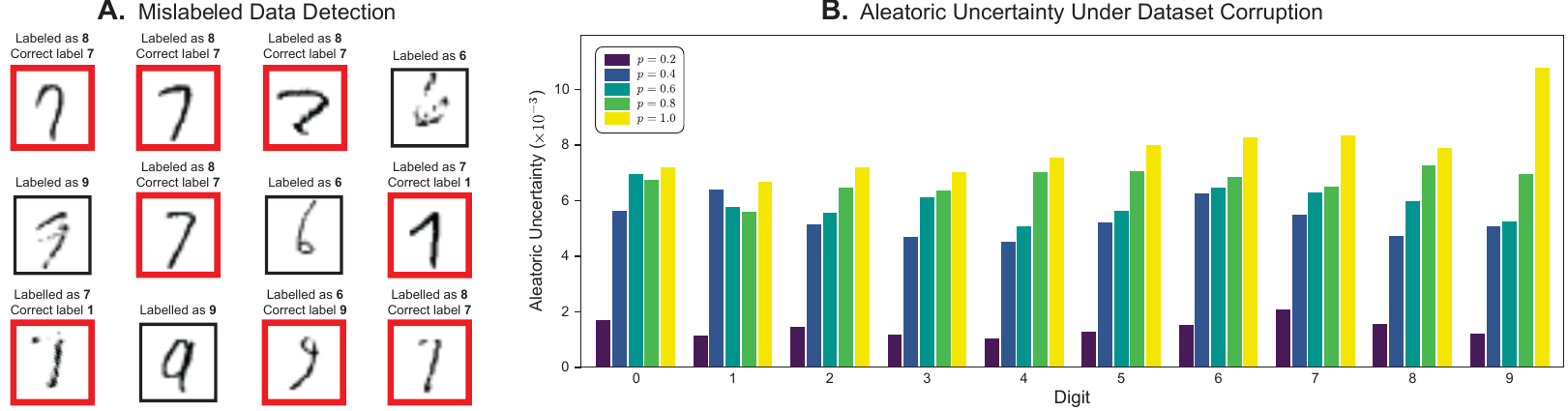}
\caption{\textbf{Mislabeled Examples in the MNIST dataset} (A) If we purposefully inject label noise into the MNIST dataset by labeling 20\% of the 7s in the dataset as 8, the mislabeled items have the highest aleatoric uncertainty. We also find a naturally mislabeled sample in the dataset. (B) As the percentage of mislabeled items increases, the average measured aleatoric uncertainty per class also increases.}
\label{fig:aleatoric-mnist}
\end{figure*}

Another application of \capsa~ is cleaning mislabeled or noisy datasets. We previously described \capsa's ability to find noisy labels with high accuracy in \sect{res:aleatoric}. In the following experiment, we replaced a random collection of the 7s in the MNIST dataset with 8s. As shown in \fig{fig:aleatoric-mnist}(A), the samples with high aleatoric uncertainty are dominated by the mislabeled examples, and also include a naturally mislabeled sample. 
We further test \capsa's sensitivity to mislabeled datasets by artificially corrupting our labels with varying levels of probability $p$. In \fig{fig:aleatoric-mnist}B, as $p$ increases, the average aleatoric uncertainty per class also increases. These experiments highlight \capsa's capability to serve as the backbone of a dataset quality controller and cleaner, due to its high-fidelity aleatoric noise detection.

\subsection{Anomaly and Adversarial Noise Detection}

Another application of the “uncertainty estimation” functionality provided by \capsa~ is anomaly detection. The core idea behind this approach is that a model's epistemic uncertainty on out-of-distribution (OOD) data is naturally higher than the same model’s epistemic uncertainty on in-distribution (ID) data. Thus, given a risk aware model, we visualize density histograms of per image uncertainty estimates provided by a model on both ID (unseen test-set for NYU Depth V2 dataset) and OOD data (ApolloScapes) (see \fig{fig:depth}E). At this point, OOD detection is possible by a simple thresholding. We use AUC-ROC to quantitatively assess the separation of the two density histograms, a higher AUC indicates a better quality of the separation (see \fig{fig:depth}D) .

\begin{figure}[t!]
\centering
\includegraphics[width=\linewidth]{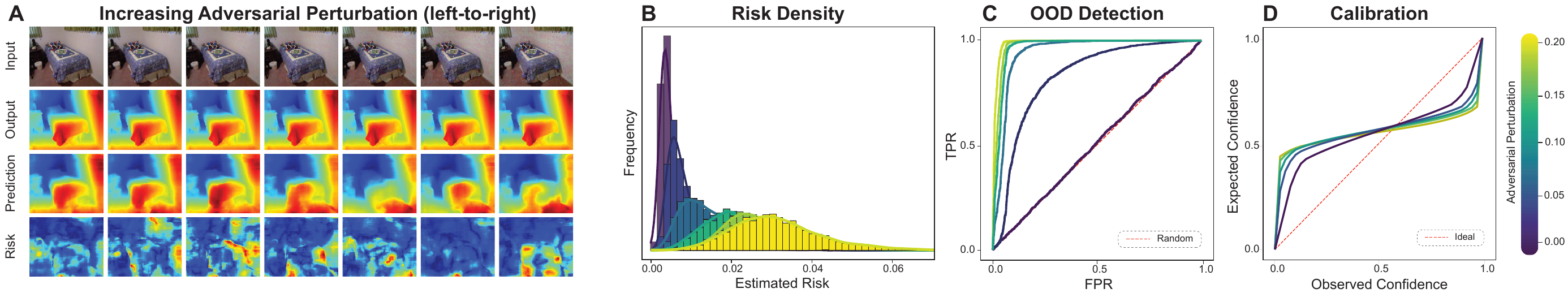}
\caption{\textbf{Robustness under adversarial noise} Across increasing levels of adversarial perturbations: (A) Pixel-wise depth predictions and uncertainty visualizations, (B) Density histograms of per image uncertainty, (C) OOD detection assessed via AUC-ROC, (D) Calibration curves.}
\label{fig:adversarial}
\end{figure}

It is critical for a model to recognize that it is presented with an unreasonable input (e.g., OOD). This capability could be used for autonomous vehicles that yield control to humans when model performance is expected to be poor. For example, in \fig{fig:adversarial}A we see that depth estimates degrade as images drift from the distribution. We are able to detect these shifts and can use this information to avoid incorrect predictions. 

Further, the approach described above could be used to detect adversarial attacks (perturbations). In \fig{fig:adversarial}A we see that even though the perturbed images are not immediately distinguishable to a human eye, the method described above successfully detects the altered input images.

Another way of interpreting the adversarial perturbations is as a way of gradually turning ID datapoints into OOD. Such a granular control allows for enhanced model introspection. In \fig{fig:adversarial} we see that as the epsilon of the perturbation increases, the density histograms of per image uncertainty estimates on both the ID and perturbed images become more disentangled (B) and thus the quality of separation increases (C).

\section{Conclusions}

In this paper, we present a unified, model-agnostic framework for risk estimation, which allows for seamless and efficient integration of uncertainty estimates in a couple lines of code. Our approach opens new avenues for greater reproducibility and benchmarking of risk and uncertainty estimation methods. We validate the scalability and the convenience of the framework on a variety of datasets. We showcase how our method can compose different algorithms together to quantify different risk metrics efficiently in parallel. We demonstrate how the obtained uncertainty estimates can be used for downstream tasks. We further show how the framework yields interpretable risk estimation results that can provide a deeper insight into decision boundaries of NNs. We refer readers 
\texttt{{Capsa Pro}} \citep{capsa-pro} for details regarding our comprehensive implementation of the functionality described in this publication. \Capsa~ has the goal of accelerating and unifying advances in the areas of uncertainty estimation and trustworthy AI. In the future, we plan to extend our approach to other data modalities including irregular types (graphs) and temporal data (sequences), as well as to support other model types and other risk metrics.




\bibliography{refs,refs_evidence}
\bibliographystyle{iclr2023_conference}

\clearpage
\appendix
\section{Appendix}

\begin{figure*}[h!]
 \includegraphics[width=.48\linewidth]{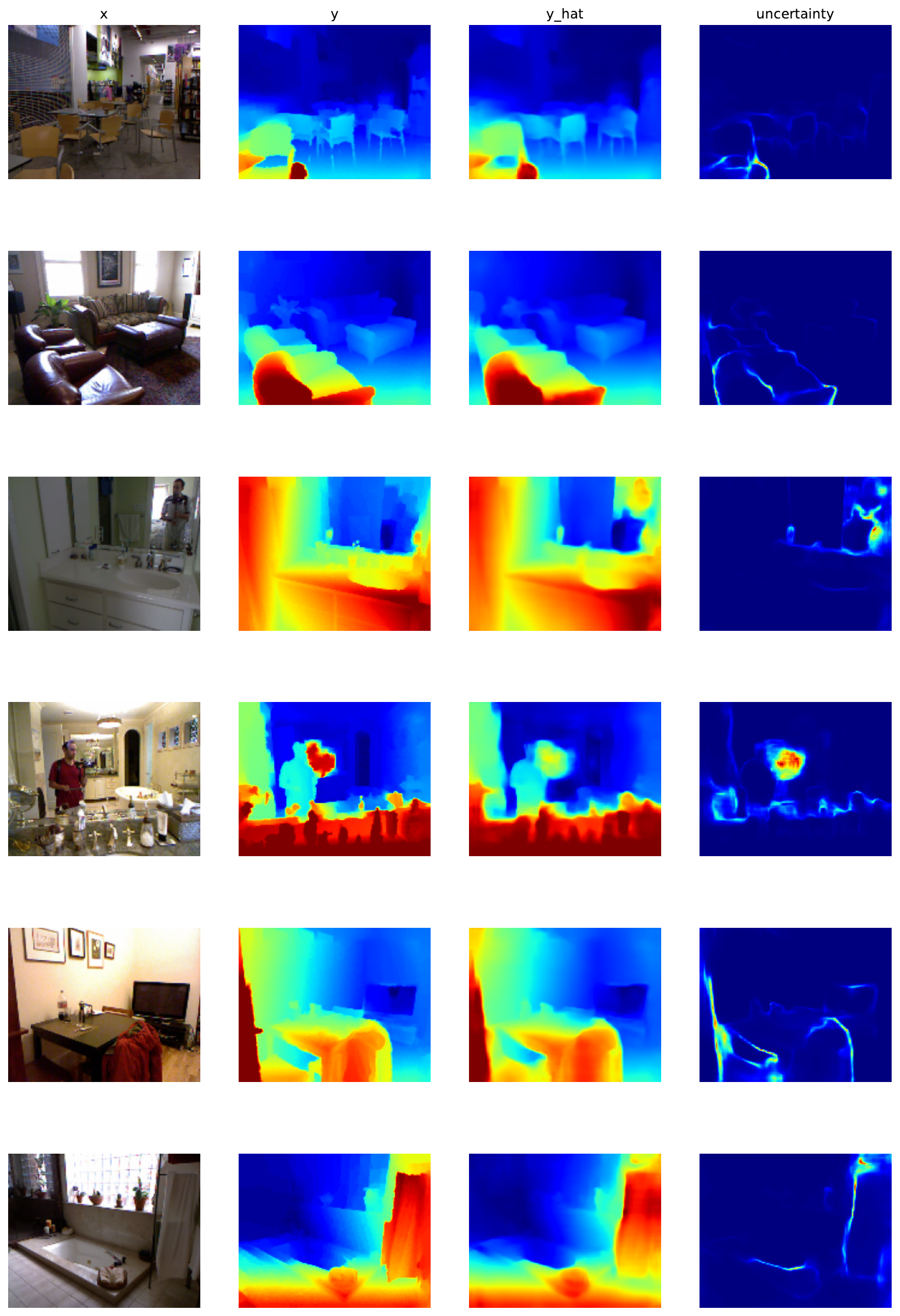} \hfill
 \includegraphics[width=.48\linewidth]{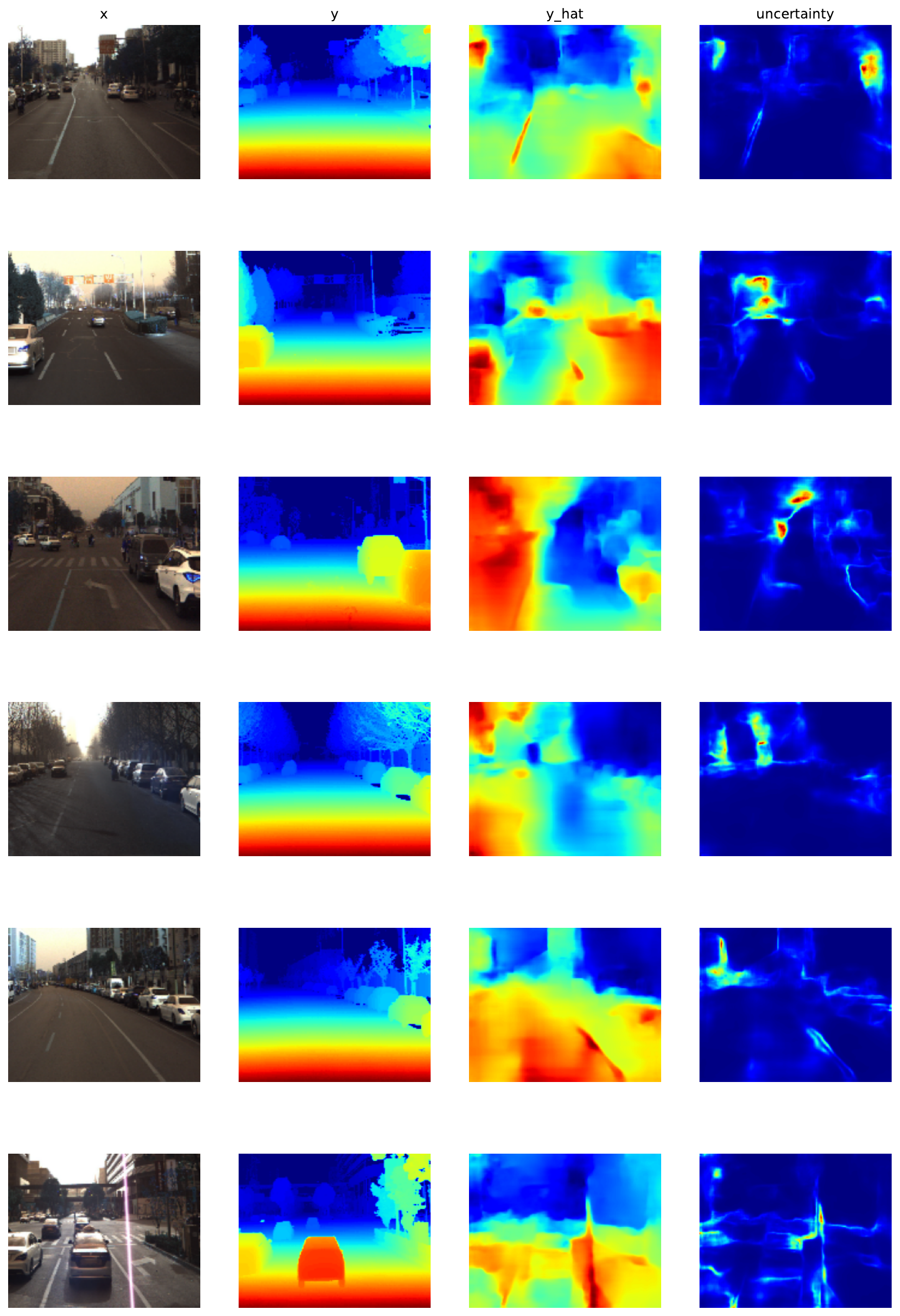} \\
\caption{MVE Wrapper}
\label{fig:depth_mve}
\end{figure*}

We see that when we wrap the model with an aleatoric method in \fig{fig:depth_mve}, we can successfully detect label noise or mislabeled data. E.g., in the 4th row on the left we see that the ground truth label has a mislabeled blob of pixels near the right shoulder of a person. The wrapped model is able to detect this and selectively assign high aleatoric uncertainty to this region while leaving the correctly labeled parts of the image ``untouched''.

\begin{table}[h!]
\resizebox{1\textwidth}{!}{

\begin{tabular}{@{}lllll@{}}
\toprule
               & Model Modifications                                       & New Models                    & Loss Changes                                                                 & Uncertainty Estimation                  \\ \midrule
Ensemble       & None                                                      & Train N - 1 additional models & N loss functions and N optimizers                                            & Uncertainty Estimation (Classification) \\
MVE            & Adding layers for sigma                                   & None                          & Train using NLL for regression, train on perturbed logits for classification & Predicted variance                      \\
VAE            & None                                                      & Adding decoder model          & MSE + KL-loss                                                                & MSE(x, \textbackslash{}hat\{x\})        \\
Dropout        & Dropout layers after fully connected/convolutional layers & None                          & None                                                                         & Variance of T forward passes            \\
Histogram Bias & Calculate histogram after every batch                     & None                          & None                                                                         & Joint probability of sample features    \\ \bottomrule
\end{tabular}
}
\end{table}

\begin{table}[h!]
\centering
\caption{\textbf{}{VAE + Dropout composability} Results of composability experiments on UCI datasets. The NLL reduces drastically for most datasets between pure VAE and VAE + Dropout, and the RMSE remains competitive, showing that composability improves uncertainty estimation quality.}
\label{tab:vae-dropout}
\resizebox{0.4\textwidth}{!}{
\begin{tabular}{@{}lcc@{}}
\toprule
            & \textbf{RMSE} & \textbf{NLL}   \\ \midrule
Boston      & 2.278 ± 0.113 & 2.292 ± 0.056  \\
Power-Plant & 4.323 ± 0.017 & 2.891 ± 0.010  \\
Yacht       & 1.630 ± 0.241 & 2.081± 0.072   \\
Concrete    & 6.489 ± 0.067 & 3.396 ± 0.036  \\
Naval       & 0.000 ± 0.000 & -2.305 ± 0.144 \\
Energy      & 1.653 ± 0.200 & 2.027 ± 0.068  \\
Kin8nm      & 0.087 ± 0.000 & -1.000 ± 0.036 \\
Protein     & 4.469 ± 0.025 & 3.156 ± 0.189  \\ \bottomrule
\end{tabular}
}
\end{table}

\begin{table*}[h!]
\centering
\caption{Training times in seconds of different metrics and composability schemes on real-world regression datasets. }
\label{tab:training-times}
\resizebox{0.7\textwidth}{!}{
\begin{tabular}{@{}lcccc@{}}
\toprule
            & \textbf{Ensembles} & \textbf{Dropout} & \textbf{VAE} & \textbf{VAE + Dropout} \\ \midrule
Boston      & 11.4 ± 1.1         & 5.9 ± 1.1        & 7.8 ± 0.1    & 7.8 ± 0.2              \\
Power-Plant & 80.2 ± 1.2         & 31.3 ± 0.6       & 51.1 ± 1.3   & 54.6 ± 1.0             \\
Yacht       & 32.4 ± 0.7         & 14.01 ± 0.1      & 19.6 ± 0.5   & 22.1 ± 0.3             \\
Concrete    & 104.8 ± 2.6        & 41.4 ± 0.6       & 57.8 ± 1.5   & 61.3 ± 0.9             \\
Naval       & 58.3 ± 0.4         & 20.0 ± 0.3       & 33.2 ± 0.5   & 35.3 ± 0.1             \\
Energy      & 87.1 ± 0.8         & 31.7 ± 0.7       & 43.9 ± 1.3   & 46.2 ± 1.0             \\
Kin8nm      & 857.0 ± 57.0       & 310.6 ± 4.3      & 440.3 ± 5.8  & 457.4 ± 11.3           \\
Protein     & 96.2 ± 0.8         & 35.7 ± 0.6       & 59.6 ± 1.0   & 64.5 ± 1.0             \\ \bottomrule
\end{tabular}
}
\end{table*}


\begin{figure*}[h!]
 \includegraphics[width=.48\linewidth]{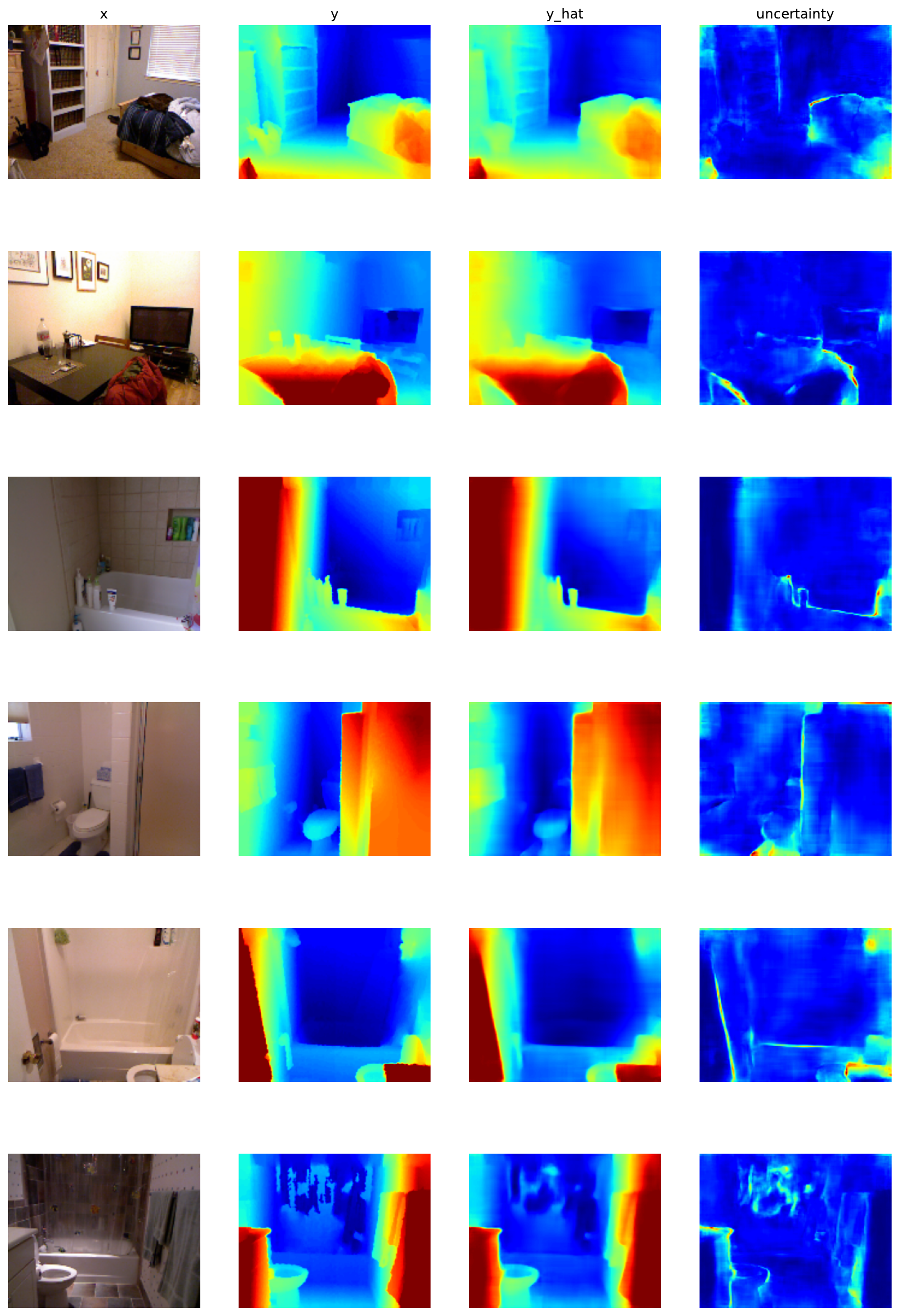} \hfill
 \includegraphics[width=.48\linewidth]{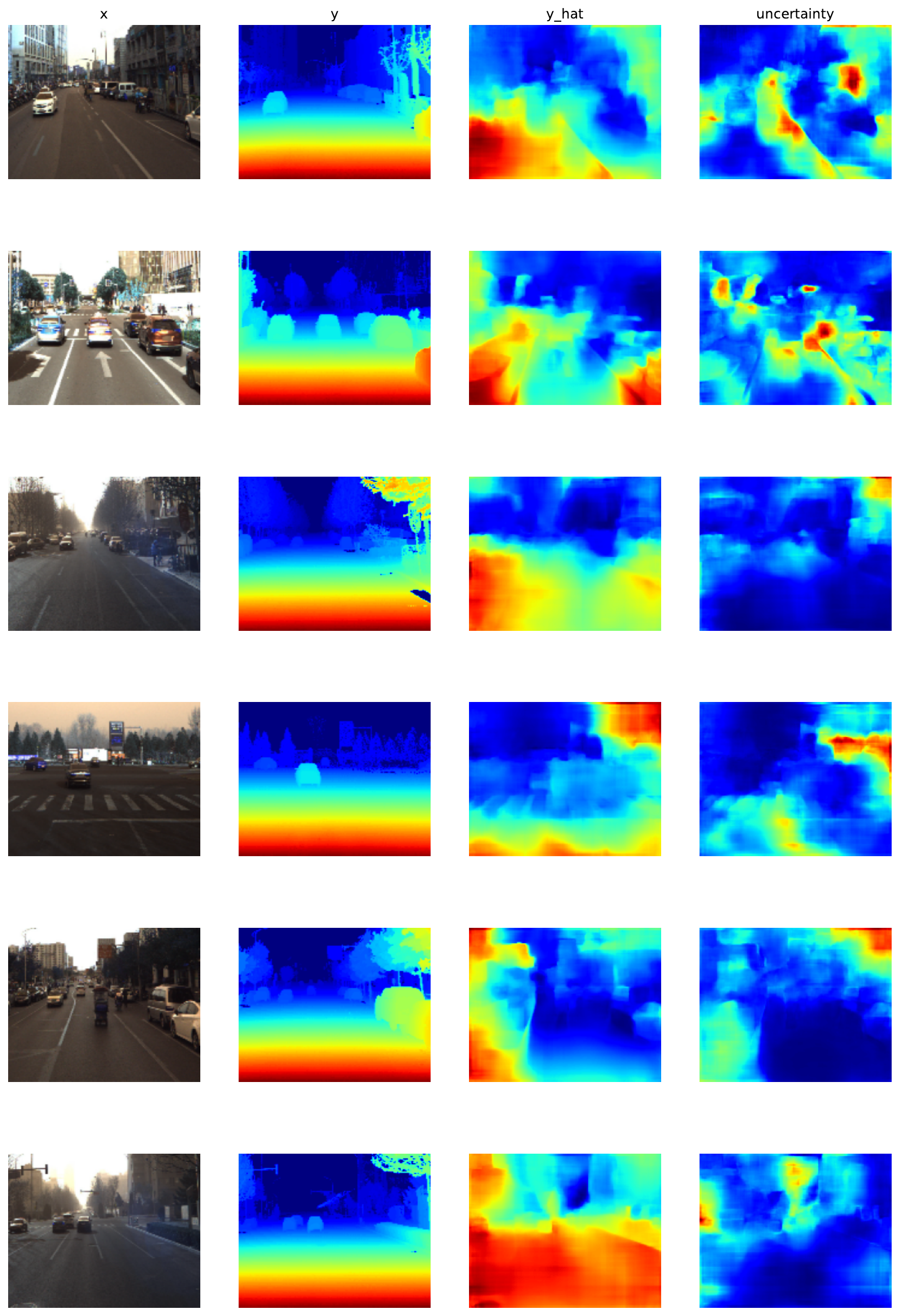} \\
\caption{Dropout Wrapper}
\label{fig:depth_dropout}
\end{figure*}

\begin{figure*}[h!]
 \includegraphics[width=.48\linewidth]{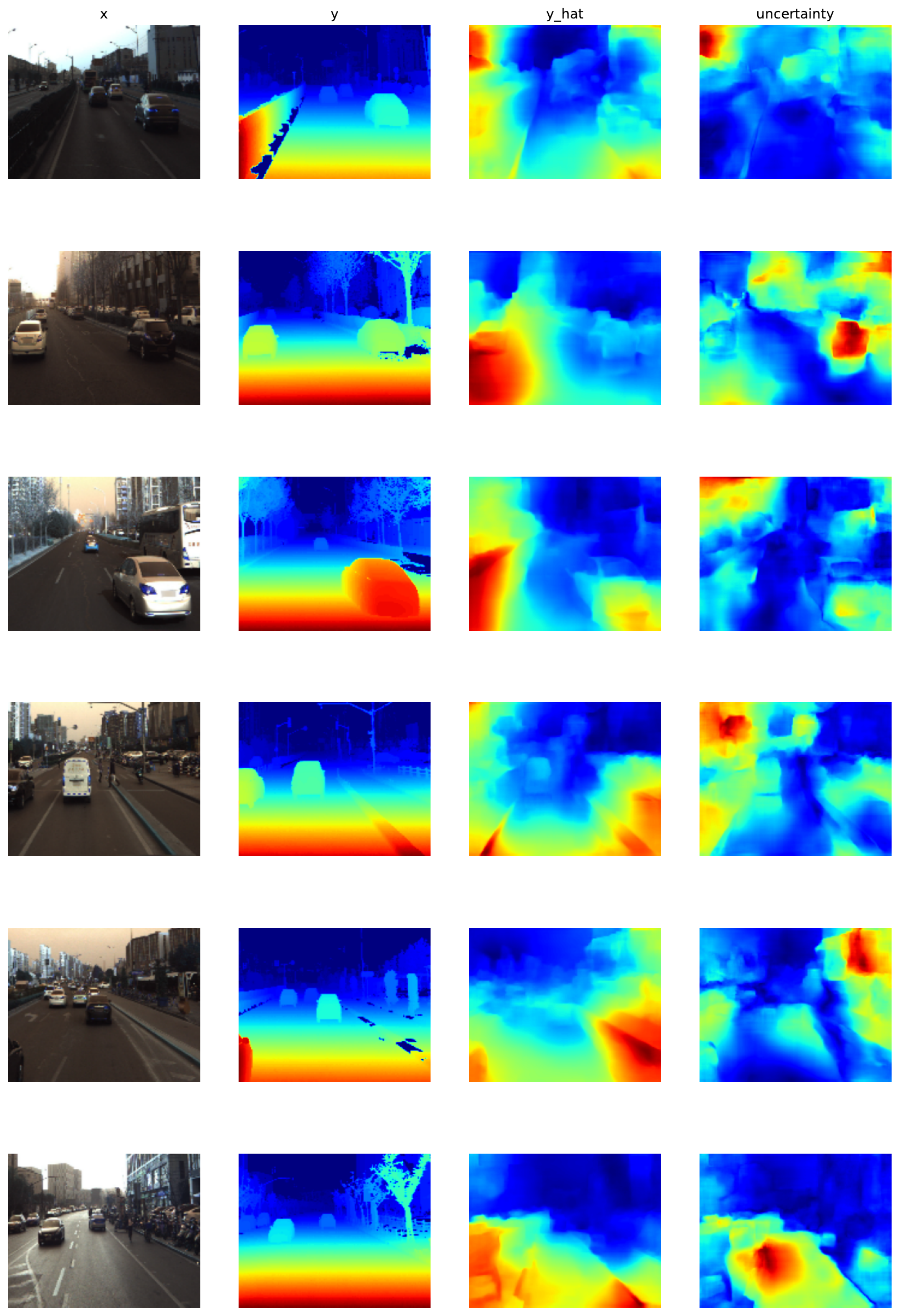} \hfill
 \includegraphics[width=.48\linewidth]{figures/depth/ensembles_ood.pdf} \\
\caption{Ensembles Wrapper}
\label{fig:depth_ensemble}
\end{figure*}

\begin{figure*}[h!]
 \includegraphics[width=.39\linewidth]{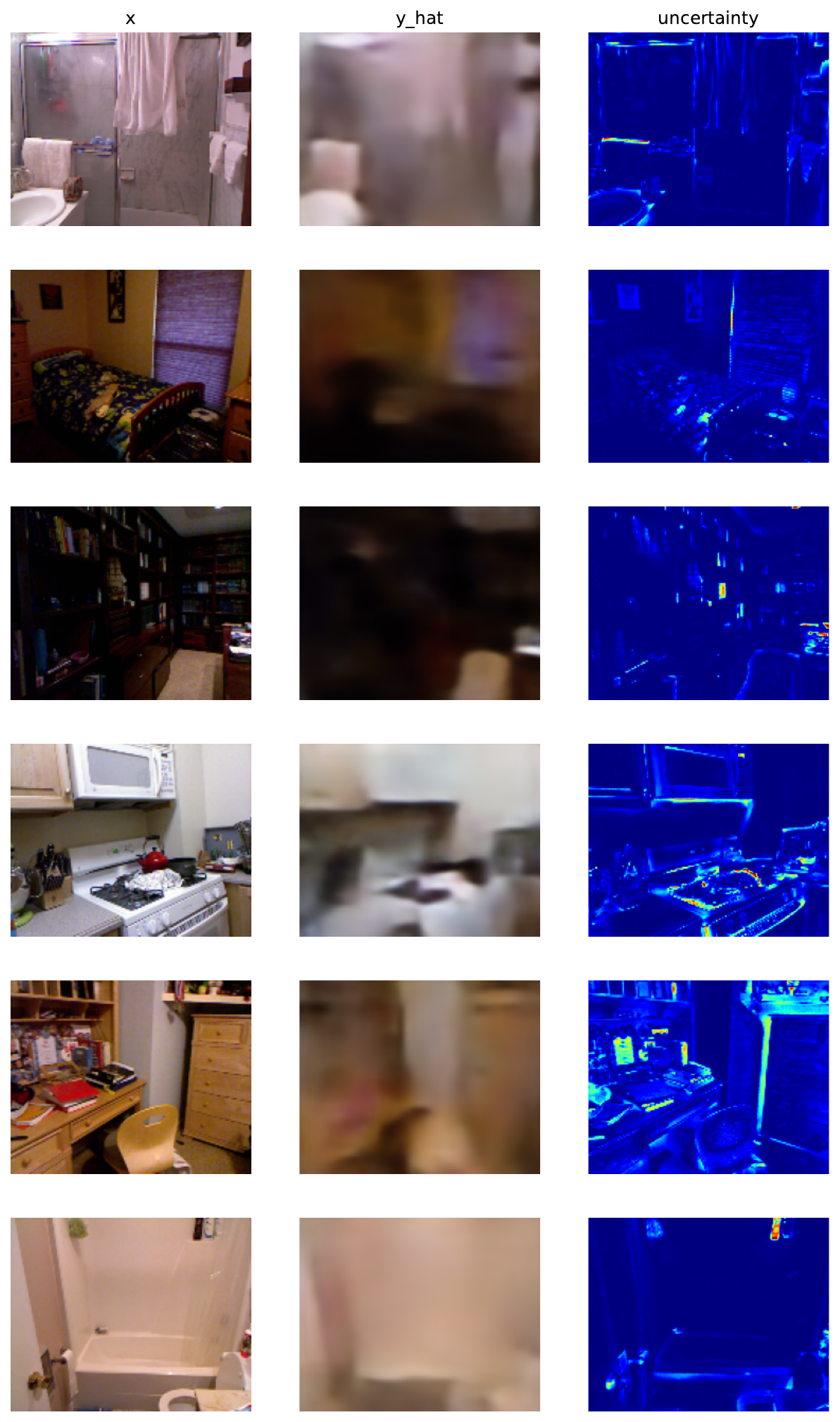} \hfill
 \includegraphics[width=.39\linewidth]{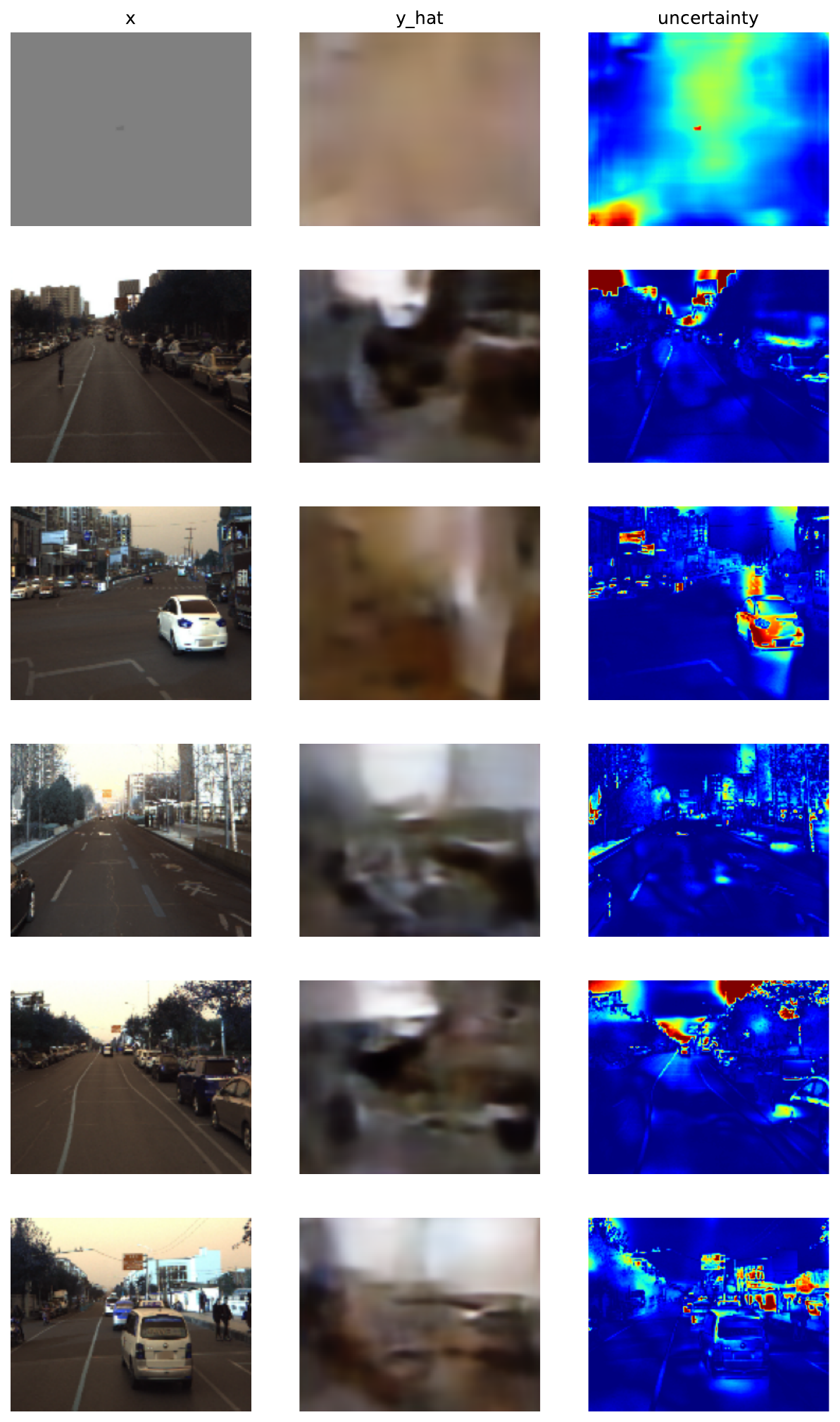} \\
\caption{VAE Wrapper}
\label{fig:depth_vae}
\end{figure*}

\end{document}

%% file: math_commands.tex
\usepackage{amsmath,amsfonts,bm}

\newcommand{\sect}[1]{Sec.~\ref{#1}}

\newcommand{\fig}[1]{Fig.~\ref{#1}}

\newcommand{\tab}[1]{Tab.~\ref{#1}}

\newcommand{\alg}[1]{Alg.~\ref{#1}}

\def\eqref#1{equation~\ref{#1}}

\def\1{\bm{1}}

\DeclareMathAlphabet{\mathsfit}{\encodingdefault}{\sfdefault}{m}{sl}
\SetMathAlphabet{\mathsfit}{bold}{\encodingdefault}{\sfdefault}{bx}{n}



%% file: uci_benchmark_table.tex
\begin{table*}[t!]
\caption{\textbf{Regression benchmarking on the UCI datasets}}
\label{tab:uci-benchmarks}
\resizebox{1\textwidth}{!}{

\begin{tabular}{@{}l|ccc|ccc@{}}
\toprule
            & \multicolumn{3}{c|}{\textbf{RMSE}}                    & \multicolumn{3}{c}{\textbf{NLL}}                      \\
            & Dropout         & VAE              & Ensemble         & Dropout         & VAE              & Ensemble         \\ \midrule
Boston      & 2.449 +/- 0.134 & 2.323 +/- 0.117  & 2.589 +/- 0.113  & 2.282 +/- 0.03  & 2.497 +/- 0.047  & 2.253 +/- 0.056  \\
Power-Plant & 4.327 +/- 0.030 & 4.286 +/- 0.0120 & 4.221 +/- 0.028  & 2.892 +/- 0.00  & 2.964 +/- 0.00   & 2.841 +/- 0.005  \\
Yacht       & 1.540 +/- 0.133 & 1.418 +/- 0.222  & 1.393 +/- 0.0965 & 2.399 +/- 0.03  & 2.637 +/- 0.131  & 1.035 +/- 0.116  \\
Concrete    & 6.628 +/- 0.286 & 6.382 +/- 0.101  & 6.456 +/- 0.846  & 3.427 +/- 0.042 & 3.361 +/- 0.016  & 3.139 +/- 0.115  \\
Naval       & 0.00 +/- 0.000  & 0.00 +/- 0.000   & 0.00 +/- 0.00    & 1.453 +/- 0.667 & -2.482 +/- 0.229 & -3.542 +/- 0.015 \\
Energy      & 1.661 +/- 0.090 & 1.377 +/- 0.091  & 1.349 +/- 0.175  & 2.120 +/- 0.022 & 1.999 +/- 0.113  & 1.395 +/- 0.066  \\
Kin8nm      & 0.088 +/- 0.001 & 0.0826 +/- 0.001 & 0.072 +/- 0.000  & -0.972 +/- 0.01 & -0.913 +/- 0.00  & -1.26 +/- 0.008  \\
Protein     & 4.559 +/- 0.031 & 4.361 +/- 0.0156 & 4.295 +/- 0.029  & 4.452 +/- 0.012 & 3.345 +/- 0.011  & 2.723 +/- 0.023  \\ \bottomrule
\end{tabular}

}
\end{table*}

%% file: depth_benchmark_table.tex
\begin{wraptable}{l}{0.59\textwidth}

\centering
\vspace{-15pt}

\caption{\textbf{Depth regression results.} VAE + dropout outperforms all other epistemic methods and is more efficient.}
\label{tab:depth-benchmark}
\resizebox{0.59\textwidth}{!}{
\begin{tabular}{@{}cccc@{}}
\toprule
              & \textbf{Test Loss}      & \textbf{NLL}             & \textbf{OOD AUC}         \\ \midrule
Base          & 0.0027 ± 0.0002         & –                        & –                        \\ \midrule
VAE           & 0.0027 ± 0.0001         & –                        & 0.8855 ± 0.0361          \\
Dropout       & 0.0027 ± 0.0001         & 0.1397 ± 0.0123          & 0.9986 ± 0.0026          \\
Ensembles     & \textbf{0.0023 ± 7e-05} & 0.0613 ± 0.0217          & 0.9989 ± 0.0018          \\
MVE           & 0.0036 ± 0.0010         & \textbf{0.0532 ± 0.0224} & 0.9798 ± 0.0118          \\ \midrule
Dropout + MVE & 0.0027 ± 0.0001         & 0.1291 ± 0.0146          & 0.9986 ± 0.0026          \\
VAE + Dropout & 0.0027 ± 0.0001         & \textbf{0.0932 ± 0.0201} & \textbf{0.9988 ± 0.0024} \\
VAE + MVE     & 0.0034 ± 0.0012         & 0.1744 ± 0.0156          & 0.9823 ± 0.0102          \\ \bottomrule
\end{tabular}

}
\end{wraptable}

%% file: iclr2023_conference.bbl
\begin{thebibliography}{41}
\providecommand{\natexlab}[1]{#1}
\providecommand{\url}[1]{\texttt{#1}}
\expandafter\ifx\csname urlstyle\endcsname\relax
  \providecommand{\doi}[1]{doi: #1}\else
  \providecommand{\doi}{doi: \begingroup \urlstyle{rm}\Url}\fi

\bibitem[Amini et~al.(2018)Amini, Schwarting, Rosman, Araki, Karaman, and
  Rus]{amini2018variational}
Alexander Amini, Wilko Schwarting, Guy Rosman, Brandon Araki, Sertac Karaman,
  and Daniela Rus.
\newblock Variational autoencoder for end-to-end control of autonomous driving
  with novelty detection and training de-biasing.
\newblock In \emph{2018 IEEE/RSJ International Conference on Intelligent Robots
  and Systems (IROS)}, pp.\  568--575. IEEE, 2018.

\bibitem[Amini et~al.(2019)Amini, Soleimany, Schwarting, Bhatia, and
  Rus]{amini2019uncovering}
Alexander Amini, Ava~P Soleimany, Wilko Schwarting, Sangeeta~N Bhatia, and
  Daniela Rus.
\newblock Uncovering and mitigating algorithmic bias through learned latent
  structure.
\newblock In \emph{Proceedings of the 2019 AAAI/ACM Conference on AI, Ethics,
  and Society}, pp.\  289--295, 2019.

\bibitem[Amini et~al.(2020)Amini, Schwarting, Soleimany, and
  Rus]{amini2020deep}
Alexander Amini, Wilko Schwarting, Ava Soleimany, and Daniela Rus.
\newblock Deep evidential regression.
\newblock \emph{Advances in Neural Information Processing Systems},
  33:\penalty0 14927--14937, 2020.

\bibitem[Amini et~al.(2023)]{capsa-pro}
Alexander Amini et~al.
\newblock Capsa pro, 2023.
\newblock URL \url{https://themisai.io/capsa-pro}.

\bibitem[Beigman \& Klebanov(2009)Beigman and Klebanov]{beigman2009learning}
Eyal Beigman and Beata~Beigman Klebanov.
\newblock Learning with annotation noise.
\newblock In \emph{Proceedings of the Joint Conference of the 47th Annual
  Meeting of the ACL and the 4th International Joint Conference on Natural
  Language Processing of the AFNLP}, pp.\  280--287, 2009.

\bibitem[Bingham et~al.(2019)Bingham, Chen, Jankowiak, Obermeyer, Pradhan,
  Karaletsos, Singh, Szerlip, Horsfall, and Goodman]{bingham2019pyro}
Eli Bingham, Jonathan~P Chen, Martin Jankowiak, Fritz Obermeyer, Neeraj
  Pradhan, Theofanis Karaletsos, Rohit Singh, Paul Szerlip, Paul Horsfall, and
  Noah~D Goodman.
\newblock Pyro: Deep universal probabilistic programming.
\newblock \emph{The Journal of Machine Learning Research}, 20\penalty0
  (1):\penalty0 973--978, 2019.

\bibitem[Blundell et~al.(2015)Blundell, Cornebise, Kavukcuoglu, and
  Wierstra]{blundell2015weight}
Charles Blundell, Julien Cornebise, Koray Kavukcuoglu, and Daan Wierstra.
\newblock Weight uncertainty in neural networks.
\newblock \emph{arXiv preprint arXiv:1505.05424}, 2015.

\bibitem[Bojarski et~al.(2016)Bojarski, Del~Testa, Dworakowski, Firner, Flepp,
  Goyal, Jackel, Monfort, Muller, Zhang, et~al.]{bojarski2016end}
Mariusz Bojarski, Davide Del~Testa, Daniel Dworakowski, Bernhard Firner, Beat
  Flepp, Prasoon Goyal, Lawrence~D Jackel, Mathew Monfort, Urs Muller, Jiakai
  Zhang, et~al.
\newblock End to end learning for self-driving cars.
\newblock \emph{arXiv preprint arXiv:1604.07316}, 2016.

\bibitem[Bolukbasi et~al.(2016)Bolukbasi, Chang, Zou, Saligrama, and
  Kalai]{bolukbasi2016man}
Tolga Bolukbasi, Kai-Wei Chang, James~Y Zou, Venkatesh Saligrama, and Adam~T
  Kalai.
\newblock Man is to computer programmer as woman is to homemaker? debiasing
  word embeddings.
\newblock \emph{Advances in neural information processing systems}, 29, 2016.

\bibitem[Buda et~al.(2018)Buda, Maki, and Mazurowski]{buda2018systematic}
Mateusz Buda, Atsuto Maki, and Maciej~A Mazurowski.
\newblock A systematic study of the class imbalance problem in convolutional
  neural networks.
\newblock \emph{Neural networks}, 106:\penalty0 249--259, 2018.

\bibitem[Buolamwini \& Gebru(2018)Buolamwini and Gebru]{buolamwini2018gender}
Joy Buolamwini and Timnit Gebru.
\newblock Gender shades: Intersectional accuracy disparities in commercial
  gender classification.
\newblock In \emph{Conference on fairness, accountability and transparency},
  pp.\  77--91. PMLR, 2018.

\bibitem[Caliskan et~al.(2017)Caliskan, Bryson, and
  Narayanan]{caliskan2017semantics}
Aylin Caliskan, Joanna~J Bryson, and Arvind Narayanan.
\newblock Semantics derived automatically from language corpora contain
  human-like biases.
\newblock \emph{Science}, 356\penalty0 (6334):\penalty0 183--186, 2017.

\bibitem[Chen et~al.(2018)Chen, Johansson, and Sontag]{chen2018my}
Irene Chen, Fredrik~D Johansson, and David Sontag.
\newblock Why is my classifier discriminatory?
\newblock \emph{Advances in neural information processing systems}, 31, 2018.

\bibitem[Ching et~al.(2018)Ching, Himmelstein, Beaulieu-Jones, Kalinin, Do,
  Way, Ferrero, Agapow, Zietz, Hoffman, et~al.]{ching2018opportunities}
Travers Ching, Daniel~S Himmelstein, Brett~K Beaulieu-Jones, Alexandr~A
  Kalinin, Brian~T Do, Gregory~P Way, Enrico Ferrero, Paul-Michael Agapow,
  Michael Zietz, Michael~M Hoffman, et~al.
\newblock Opportunities and obstacles for deep learning in biology and
  medicine.
\newblock \emph{Journal of The Royal Society Interface}, 15\penalty0
  (141):\penalty0 20170387, 2018.

\bibitem[Codevilla et~al.(2018)Codevilla, M{\"u}ller, L{\'o}pez, Koltun, and
  Dosovitskiy]{codevilla2018end}
Felipe Codevilla, Matthias M{\"u}ller, Antonio L{\'o}pez, Vladlen Koltun, and
  Alexey Dosovitskiy.
\newblock End-to-end driving via conditional imitation learning.
\newblock In \emph{2018 IEEE international conference on robotics and
  automation (ICRA)}, pp.\  4693--4700. IEEE, 2018.

\bibitem[Dillon et~al.(2017)Dillon, Langmore, Tran, Brevdo, Vasudevan, Moore,
  Patton, Alemi, Hoffman, and Saurous]{dillon2017tensorflow}
Joshua~V Dillon, Ian Langmore, Dustin Tran, Eugene Brevdo, Srinivas Vasudevan,
  Dave Moore, Brian Patton, Alex Alemi, Matt Hoffman, and Rif~A Saurous.
\newblock Tensorflow distributions.
\newblock \emph{arXiv preprint arXiv:1711.10604}, 2017.

\bibitem[Gal \& Ghahramani(2016)Gal and Ghahramani]{gal2016dropout}
Yarin Gal and Zoubin Ghahramani.
\newblock Dropout as a bayesian approximation: Representing model uncertainty
  in deep learning.
\newblock In \emph{international conference on machine learning}, pp.\
  1050--1059, 2016.

\bibitem[Gilitschenski et~al.(2019)Gilitschenski, Sahoo, Schwarting, Amini,
  Karaman, and Rus]{gilitschenski2019deep}
Igor Gilitschenski, Roshni Sahoo, Wilko Schwarting, Alexander Amini, Sertac
  Karaman, and Daniela Rus.
\newblock Deep orientation uncertainty learning based on a bingham loss.
\newblock In \emph{International Conference on Learning Representations}, 2019.

\bibitem[Hawke et~al.(2020)Hawke, Shen, Gurau, Sharma, Reda, Nikolov, Mazur,
  Micklethwaite, Griffiths, Shah, et~al.]{hawke2020urban}
Jeffrey Hawke, Richard Shen, Corina Gurau, Siddharth Sharma, Daniele Reda,
  Nikolay Nikolov, Przemys{\l}aw Mazur, Sean Micklethwaite, Nicolas Griffiths,
  Amar Shah, et~al.
\newblock Urban driving with conditional imitation learning.
\newblock In \emph{2020 IEEE International Conference on Robotics and
  Automation (ICRA)}, pp.\  251--257. IEEE, 2020.

\bibitem[He \& Garcia(2009)He and Garcia]{he2009learning}
Haibo He and Edwardo~A Garcia.
\newblock Learning from imbalanced data.
\newblock \emph{IEEE Transactions on knowledge and data engineering},
  21\penalty0 (9):\penalty0 1263--1284, 2009.

\bibitem[Kendall \& Gal(2017)Kendall and Gal]{kendall2017uncertainties}
Alex Kendall and Yarin Gal.
\newblock What uncertainties do we need in bayesian deep learning for computer
  vision?
\newblock \emph{Advances in neural information processing systems}, 30, 2017.

\bibitem[Kingma \& Welling(2013)Kingma and Welling]{kingma2013auto}
Diederik~P Kingma and Max Welling.
\newblock Auto-encoding variational bayes.
\newblock \emph{arXiv preprint arXiv:1312.6114}, 2013.

\bibitem[Kompa et~al.(2021)Kompa, Snoek, and Beam]{kompa2021second}
Benjamin Kompa, Jasper Snoek, and Andrew~L Beam.
\newblock Second opinion needed: communicating uncertainty in medical machine
  learning.
\newblock \emph{NPJ Digital Medicine}, 4\penalty0 (1):\penalty0 1--6, 2021.

\bibitem[Lakshminarayanan et~al.(2017)Lakshminarayanan, Pritzel, and
  Blundell]{lakshminarayanan2017simple}
Balaji Lakshminarayanan, Alexander Pritzel, and Charles Blundell.
\newblock Simple and scalable predictive uncertainty estimation using deep
  ensembles.
\newblock \emph{Advances in neural information processing systems}, 30, 2017.

\bibitem[Liao et~al.(2020)Liao, Lu, Zhou, Zhang, Li, and Yang]{liao2020dvi}
Miao Liao, Feixiang Lu, Dingfu Zhou, Sibo Zhang, Wei Li, and Ruigang Yang.
\newblock Dvi: Depth guided video inpainting for autonomous driving.
\newblock In \emph{European Conference on Computer Vision}, pp.\  1--17.
  Springer, 2020.

\bibitem[Liu et~al.(2015)Liu, Luo, Wang, and Tang]{liu2015faceattributes}
Ziwei Liu, Ping Luo, Xiaogang Wang, and Xiaoou Tang.
\newblock Deep learning face attributes in the wild.
\newblock In \emph{Proceedings of International Conference on Computer Vision
  (ICCV)}, December 2015.

\bibitem[Nado et~al.(2021)Nado, Band, Collier, Djolonga, Dusenberry, Farquhar,
  Feng, Filos, Havasi, Jenatton, et~al.]{nado2021uncertainty}
Zachary Nado, Neil Band, Mark Collier, Josip Djolonga, Michael~W Dusenberry,
  Sebastian Farquhar, Qixuan Feng, Angelos Filos, Marton Havasi, Rodolphe
  Jenatton, et~al.
\newblock Uncertainty baselines: Benchmarks for uncertainty \& robustness in
  deep learning.
\newblock \emph{arXiv preprint arXiv:2106.04015}, 2021.

\bibitem[Nathan~Silberman \& Fergus(2012)Nathan~Silberman and
  Fergus]{Silberman:ECCV12}
Pushmeet~Kohli Nathan~Silberman, Derek~Hoiem and Rob Fergus.
\newblock Indoor segmentation and support inference from rgbd images.
\newblock In \emph{ECCV}, 2012.

\bibitem[Nix \& Weigend(1994)Nix and Weigend]{nix1994estimating}
David~A Nix and Andreas~S Weigend.
\newblock Estimating the mean and variance of the target probability
  distribution.
\newblock In \emph{Proceedings of 1994 ieee international conference on neural
  networks (ICNN'94)}, volume~1, pp.\  55--60. IEEE, 1994.

\bibitem[Obermeyer et~al.(2019)Obermeyer, Powers, Vogeli, and
  Mullainathan]{obermeyer2019dissecting}
Ziad Obermeyer, Brian Powers, Christine Vogeli, and Sendhil Mullainathan.
\newblock Dissecting racial bias in an algorithm used to manage the health of
  populations.
\newblock \emph{Science}, 366\penalty0 (6464):\penalty0 447--453, 2019.

\bibitem[Puyol-Ant{\'o}n et~al.(2021)Puyol-Ant{\'o}n, Ruijsink, Piechnik,
  Neubauer, Petersen, Razavi, and King]{puyol2021fairness}
Esther Puyol-Ant{\'o}n, Bram Ruijsink, Stefan~K Piechnik, Stefan Neubauer,
  Steffen~E Petersen, Reza Razavi, and Andrew~P King.
\newblock Fairness in cardiac mr image analysis: an investigation of bias due
  to data imbalance in deep learning based segmentation.
\newblock In \emph{International Conference on Medical Image Computing and
  Computer-Assisted Intervention}, pp.\  413--423. Springer, 2021.

\bibitem[Rosenblatt(1956)]{rosenblatt1956remarks}
Murray Rosenblatt.
\newblock Remarks on some nonparametric estimates of a density function.
\newblock \emph{The annals of mathematical statistics}, pp.\  832--837, 1956.

\bibitem[Seyyed-Kalantari et~al.(2021)Seyyed-Kalantari, Zhang, McDermott, Chen,
  and Ghassemi]{seyyed2021underdiagnosis}
Laleh Seyyed-Kalantari, Haoran Zhang, Matthew McDermott, Irene~Y Chen, and
  Marzyeh Ghassemi.
\newblock Underdiagnosis bias of artificial intelligence algorithms applied to
  chest radiographs in under-served patient populations.
\newblock \emph{Nature medicine}, 27\penalty0 (12):\penalty0 2176--2182, 2021.

\bibitem[Shi et~al.(2017)Shi, Chen, Zhu, Sun, Luo, Gu, and
  Zhou]{shi2017zhusuan}
Jiaxin Shi, Jianfei Chen, Jun Zhu, Shengyang Sun, Yucen Luo, Yihong Gu, and
  Yuhao Zhou.
\newblock Zhusuan: A library for bayesian deep learning.
\newblock \emph{arXiv preprint arXiv:1709.05870}, 2017.

\bibitem[Soleimany et~al.(2021)Soleimany, Amini, Goldman, Rus, Bhatia, and
  Coley]{soleimany2021evidential}
Ava~P Soleimany, Alexander Amini, Samuel Goldman, Daniela Rus, Sangeeta~N
  Bhatia, and Connor~W Coley.
\newblock Evidential deep learning for guided molecular property prediction and
  discovery.
\newblock \emph{ACS central science}, 7\penalty0 (8):\penalty0 1356--1367,
  2021.

\bibitem[Srivastava et~al.(2014)Srivastava, Hinton, Krizhevsky, Sutskever, and
  Salakhutdinov]{srivastava2014dropout}
Nitish Srivastava, Geoffrey Hinton, Alex Krizhevsky, Ilya Sutskever, and Ruslan
  Salakhutdinov.
\newblock Dropout: a simple way to prevent neural networks from overfitting.
\newblock \emph{The journal of machine learning research}, 15\penalty0
  (1):\penalty0 1929--1958, 2014.

\bibitem[Topol(2019)]{topol2019high}
Eric~J Topol.
\newblock High-performance medicine: the convergence of human and artificial
  intelligence.
\newblock \emph{Nature medicine}, 25\penalty0 (1):\penalty0 44--56, 2019.

\bibitem[Tran et~al.(2016)Tran, Kucukelbir, Dieng, Rudolph, Liang, and
  Blei]{tran2016edward}
Dustin Tran, Alp Kucukelbir, Adji~B Dieng, Maja Rudolph, Dawen Liang, and
  David~M Blei.
\newblock Edward: A library for probabilistic modeling, inference, and
  criticism.
\newblock \emph{arXiv preprint arXiv:1610.09787}, 2016.

\bibitem[Tran et~al.(2022)Tran, Liu, Dusenberry, Phan, Collier, Ren, Han, Wang,
  Mariet, Hu, et~al.]{tran2022plex}
Dustin Tran, Jeremiah Liu, Michael~W Dusenberry, Du~Phan, Mark Collier, Jie
  Ren, Kehang Han, Zi~Wang, Zelda Mariet, Huiyi Hu, et~al.
\newblock Plex: Towards reliability using pretrained large model extensions.
\newblock \emph{arXiv preprint arXiv:2207.07411}, 2022.

\bibitem[Xu et~al.(2022)Xu, Ahmadi, Amini, Rus, and Lo]{xu2022identifying}
Qingyang Xu, Elaheh Ahmadi, Alexander Amini, Daniela Rus, and Andrew~W Lo.
\newblock Identifying and mitigating potential biases in predicting drug
  approvals.
\newblock \emph{Drug Safety}, 45\penalty0 (5):\penalty0 521--533, 2022.

\bibitem[Zhang et~al.(2018)Zhang, Lemoine, and Mitchell]{zhang2018mitigating}
Brian~Hu Zhang, Blake Lemoine, and Margaret Mitchell.
\newblock Mitigating unwanted biases with adversarial learning.
\newblock In \emph{Proceedings of the 2018 AAAI/ACM Conference on AI, Ethics,
  and Society}, pp.\  335--340, 2018.

\end{thebibliography}
